\documentclass[conference]{IEEEtran}
\IEEEoverridecommandlockouts
\usepackage{cite}
\usepackage{amsmath,amssymb,amsfonts}
\usepackage{algorithmic}
\usepackage{graphicx}
\usepackage{textcomp}
\usepackage{xcolor}
\def\BibTeX{{\rm B\kern-.05em{\sc i\kern-.025em b}\kern-.08em
    T\kern-.1667em\lower.7ex\hbox{E}\kern-.125emX}}

\usepackage[numbers,sort&compress]{natbib}
\usepackage{hyperref}
\usepackage{booktabs}
\usepackage{multirow}
\usepackage{makecell}

\usepackage[acronym, nopostdot, nonumberlist]{glossaries}
\usepackage{xpatch}
\makenoidxglossaries
\glsdisablehyper 
\newacronym{mot}{MOT}{Multi-Object Tracking}
\newacronym{sota}{SOTA}{state of the art}
\newacronym{dti}{(DTI}{Department of Trade
and Industry}
\newacronym{cair}{CAIR}{Center of Artificial Intelligence Research}

 \usepackage{colortbl}
 
\def\ccb{\cellcolor{blue!20}}
\def\ccB{\cellcolor{blue!20}\textbf}

\xpretocmd{\section}{\glsresetall}{}{}

\newcommand*\rot{\rotatebox{90}}

\begin{document}

% \title{Wildlife Inference Augmentation Using Multi-Object Tracking\\
\title{Multi-Object Tracking Consistently Improves Wildlife Inference \\
\thanks{Qulinda. World Wide Fund (WWF). Centre for Artificial Intelligence Research (CAIR).}
}

\author{\IEEEauthorblockN{Mufhumudzi Muthivhi}
\IEEEauthorblockA{\textit{Academy of Computer Science and Software Engineering} \\
\textit{University of Johannesburg}\\
Johannesburg, South Africa \\
mmuthivhi@uj.ac.za}
\and
\IEEEauthorblockN{Jiahao Huo}
\IEEEauthorblockA{\textit{Academy of Computer Science and Software Engineering} \\
\textit{University of Johannesburg}\\
Johannesburg, South Africa \\
216045414@student.uj.ac.za}
\and
\IEEEauthorblockN{Fredrik Gustafsson}
\IEEEauthorblockA{\textit{Department of Electrical Engineering} \\
\textit{Linköping University}\\
Linköping, Sweden \\
fredrik.gustafsson@liu.se}
\and
\IEEEauthorblockN{Terence L. van Zyl}
\IEEEauthorblockA{\textit{Academy of Computer Science and Software Engineering} \\
\textit{University of Johannesburg}\\
Johannesburg, South Africa \\
tvanzyl@uj.ac.za}
}

\maketitle

\begin{abstract}
% ---ESTABLISHING THE CONTEXT---
Camera traps have become a common tool for wildlife monitoring efforts in ecological research and biodiversity conservation. Wildlife classification models have benefited from the increase in wildlife visual data. These models achieve high accuracy on curated, high-quality datasets.
% ---FOCUS---
However, their performance remains sensitive to real-world environmental constraints. They often produce inconsistent predictions when inferring from temporally coherent sequences. The predicted label for a single individual shifts rapidly from frame to frame.
% ---STATING THE PURPOSE---
This study exploits the temporal nature of camera-trap data to augment inferred predictions from a wildlife classification model.
% ---DESCRIBING METHODOLOGY---
Specifically, we adopt several standard \gls{mot} models to link detections across consecutive frames. The curated trajectories are used to fuse the softmax class probabilities. The fused probability score produces a single consensus class label estimate that overrides misclassifications caused by noise.
% ---PRESENTING THE RESULTS---
The analysis of the experimental results shows that our proposed strategy outperforms a standalone classifier across all datasets and metrics. Specifically, the best-performing \gls{mot} models gain a weighted F1-Score of 5.1\%, 3.1\% and 2.0\% over the classifier across three \gls{mot} datasets. 
\end{abstract}


\section{Introduction}

% ---ESTABLISHING A RESEARCH TERRITORY---
Wildlife monitoring plays a crucial role in biodiversity conservation, ecosystem management, and ecological research. A growing body of research uses camera traps as the primary device for non-invasive, large-scale, and cost-effective data collection~\cite{caravaggi2017review, swanson2015snapshot}. The accumulated data allows researchers to document species presence, assess behavioural patterns, or estimate population size~\cite{saoud2024beyond}.
\begin{figure}[htp]
  \centering
  \begin{tabular}{c} % Single centered column
    \includegraphics[width=1.0\linewidth]{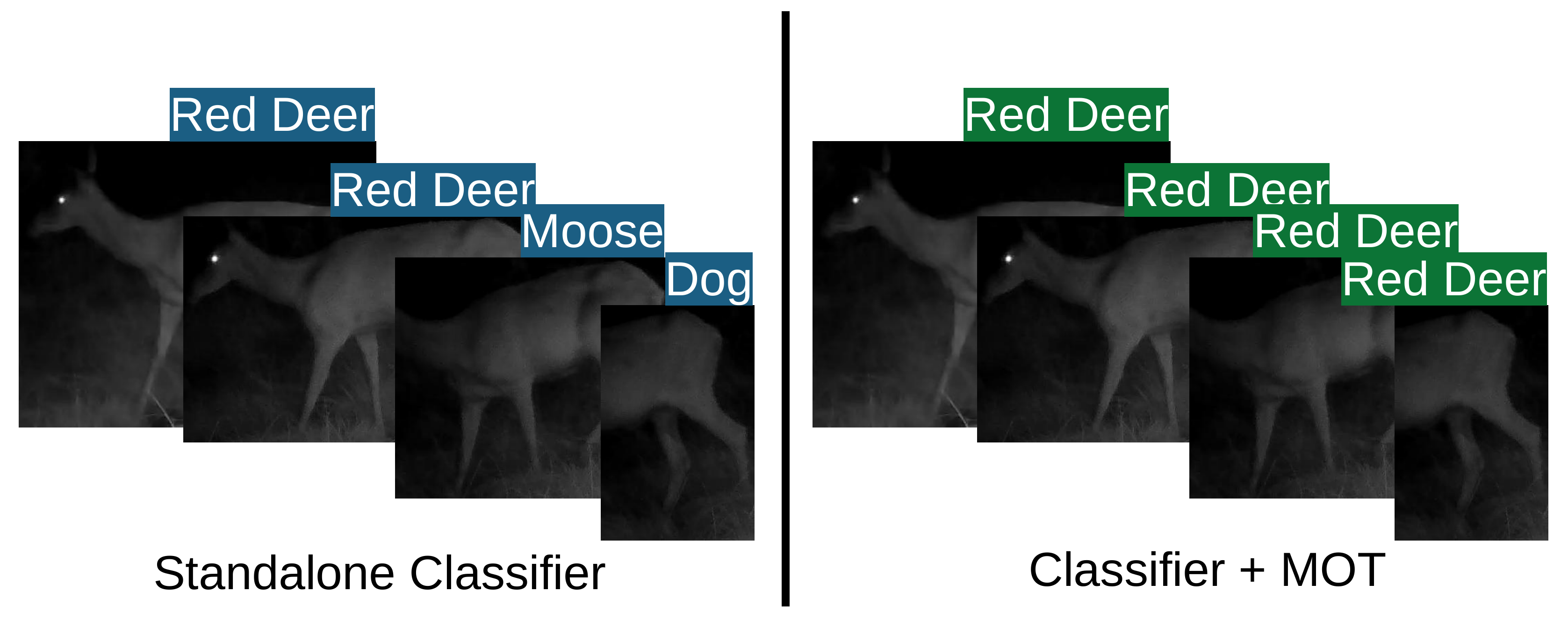} \\
    \vspace{0.5cm} % Optional: adds a bit of breathing room between the plots
    \includegraphics[width=1.0\linewidth]{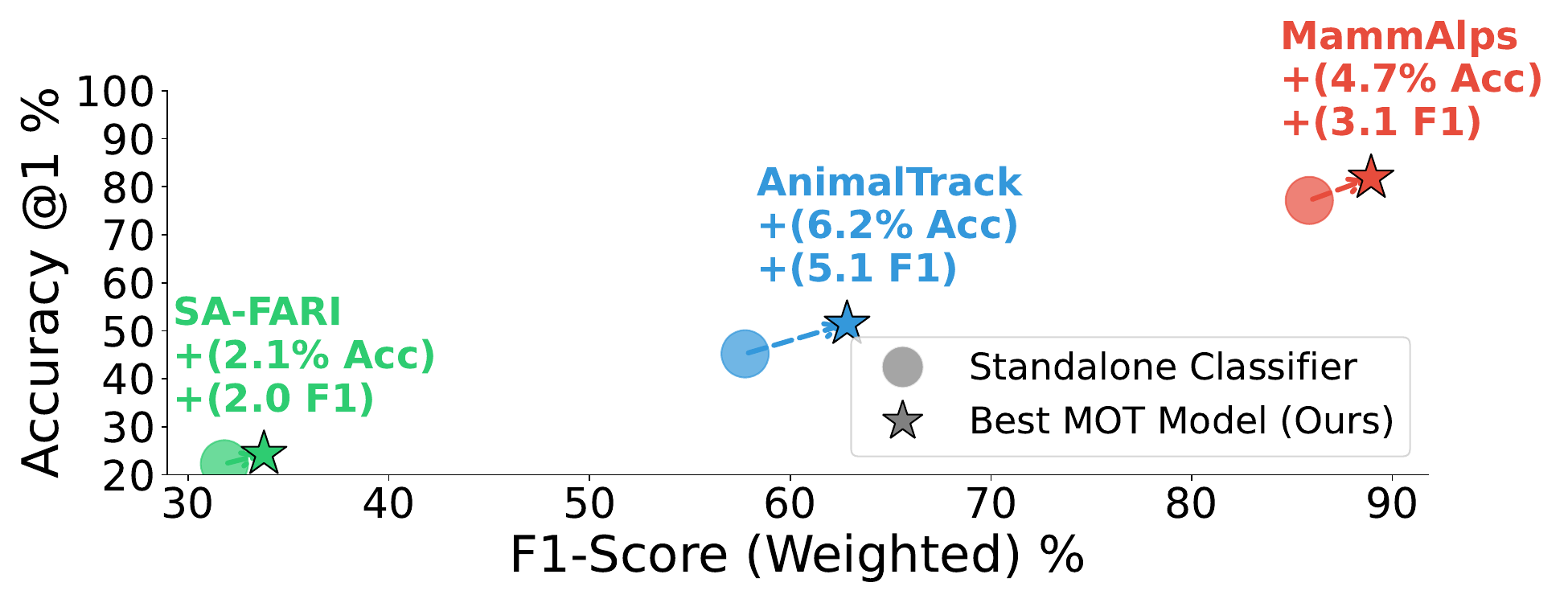} \\
  \end{tabular} 
  \caption{Comparison of model predictions (top) and overall performance (bottom). The figure shows the improvement in accuracy@1 and F1-score on recent popular wildlife benchmarks, comparing our inference augmentation to a standalone classifier. The circles represent the standalone classifier's performance, and the stars represent the improved performance achieved with Multi-Target Tracking (MOT) and our proposed class-probability fusion.}\label{figure:performance}
\end{figure}
% ---ESTABLISHING A NICHE---
Several large-scale wildlife classification models have achieved remarkable success over a large variety of animal classes~\cite{dussert2025being, willi2019identifying, muthivhi2025improving, dussert2025paying}. BioClipV2 improves species classification accuracy by 15\% above previous \gls{sota}~\cite{gu2025bioclip}. DeepFaune maintains a high classification performance across North American and European wildlife environments~\cite{clarfeld2025deepfaune, dussert2025being}. \citet{villa2017towards} uses camera-trap images from the Serengeti to train a classifier for African wildlife animals.
% SpeciesNet can classify up to 2000 animal species~\cite{gadot2024crop}.

% ---OCCUPYING THE NICHE---
However, these models typically make per-image predictions and treat sequential frames as independent observations. This approach often leads to \textit{label flickering}, where a single animal's classification toggles between different species across a sequence. Figure~\ref{figure:performance} depicts an instance of the label flickering phenomenon from a standalone classifier. Motion blur, occlusion, changing light conditions or cluttered backgrounds further exacerbate these misclassifications. 
% a) Outline your purposes and state the nature of your research
Few works exploit the temporal nature of camera trap data. \citet{liu2024improved} train a classifier using temporal metadata from camera-trap images. \citet{dussert2025being} produces a single prediction across a sequence of images taken upon triggering the camera trap~\cite{dussert2025being}. This work uses \gls{mot} across consecutive camera-trap frames to track an individual and correct predictions from a standalone classifier. 
% b) State your hypothesis or research question you seek to answer
We hypothesise that combining temporal tracking data and frame-level softmax predictions will reduce "label flickering" and improve the overall F1 score of species classification. We utilise the most informative frames within a sequence to augment the final confidence score obtained during inference. The most informative frames produce a high species prediction score due to the minimal presence of noise. Specifically, these frames have a high softmax probability score, and hence only high-confidence classifications contribute to the temporal aggregation. By tracking a detected individual over time, we filter out noise from suboptimal frames.
% c) Share your findings
Our experiments on three wildlife benchmarks show that inference-time augmentation using MOT consistently improves species classification over a per-frame standalone classifier. Across AnimalTrack, MammAlps, and SA-FARI datasets, associating detections into tracks and fusion of predictions increases macro F1 by up to 5\%, with Centroid and BotSORT providing the strongest gains while adding only a small runtime overhead relative to detection. Figure~\ref{figure:performance} depicts an increase in the accuracy and weighted F1-Score over the standalone classifier against the best performing \gls{mot} model for each dataset.
% d) Elaborate on the value of your research (Contributions)
We contribute to the existing literature by:
\begin{itemize}
    \item proposing a probability-fusion inference classification augmentation strategy using \gls{mot},
    \item conducting a comparative analysis of standard motion and appearance-based \gls{mot} frameworks,
    \item providing an ablation study on the effects of two temporal-based inference augmentation strategies,
    \item assessing the inference latency of the proposed strategy across each \gls{mot} framework, and
    \item providing a qualitative analysis of the tracked animal over a sequence of frames and depicting its augmented prediction.
\end{itemize}

\section{Related Work}

Multi-Object tracking (MOT) estimates object trajectories by localising targets each frame and maintaining consistent identities over time.
Animal MOT in the wild is difficult due to camouflage, clutter, abrupt motion, occlusion, large-scale changes, and limited labels.
Existing work broadly falls into (i) online motion-based trackers, (ii) tracking-by-detection with appearance cues, (iii) end-to-end joint detection-and-tracking models, and (iv) domain-specific adaptations and benchmarks~\cite{liu2024deep,rajagukguk2025deep}.

\subsection{Online Motion-Based Trackers.}
These methods propagate a simple kinematic state per target (often via a Kalman filter) and associate detections using geometric consistency (e.g., Intersection over Union~\cite{Rezatofighi_2018_CVPR} or Centroids~\cite{Nascimento1999AnAF}).
Online trackers are efficient and easy to deploy, but can suffer from identity switches under long occlusions or irregular motion. Simple Online and Realtime Tracking (SORT)~\cite{Bewley2016_sort} is a canonical real-time baseline that performs per-frame motion prediction with a constant-velocity Kalman filter and assigns detections to tracks using IoU-based bipartite matching (typically via the Hungarian algorithm). Observation-Centric SORT (OC-SORT)~\cite{cao2023observation} extends SORT by making association more robust when detections are missing or unreliable, improving identity stability under occlusions while preserving real-time operation~\cite{Bewley2016_sort,cao2023observation}.

\subsection{Tracking-by-Detection with Appearance Cues.}
A detector produces per-frame boxes and a tracker links them using motion plus learned embeddings (Re-Identification) to better preserve identities.
DeepSORT established a widely adopted appearance-assisted tracking-by-detection baseline by extending SORT with a learned Re-ID metric for data association, improving identity consistency when motion cues are ambiguous \cite{wojke2017simple}.
BoT-SORT further strengthens association (including camera motion handling), and ByteTrack reduces fragmentation by also linking lower-confidence detections \cite{aharon2022botsort,zhang2022bytetrack}.
This family is common in wildlife because strong detectors can be trained per domain/species while reusing generic association features~\cite{feng2022multiobject}.

\subsection{End-to-End Joint Detection and Tracking.}
End-to-end multi-object tracking (MOT) approaches integrate detection and association into a single model, often maintaining persistent ``track queries'' that are updated across frames.
TrackFormer (Tracking with Transformers) and MOTR (Multi-Object Tracking with Transformers) exemplify transformer-based designs that reduce reliance on hand-crafted matching, while GTR (Global Tracking) emphasises long-range global association \cite{meinhardt2022trackformer,zeng2022motr,}.
These methods can better exploit temporal context, but typically require more data and compute, which can be limiting for rare species.

\subsection{Wildlife-Specific Settings and Benchmarks.}
Wildlife tracking often departs from pedestrian/vehicle MOT assumptions, motivating few-shot learning, additional cues and benchmarks tailored to animals and aerial monitoring \cite{feng2022multiobject,zhang2023animaltrack}.
For long-term identity across time or cameras, MOT is frequently paired with animal Re-ID, supported by identity-annotated datasets such as WildlifeReID-10k~\cite{adam2025wildlifereid}.

\subsection{Temporal Inference}
These methods involves the use of temporal models~\cite{jin2023large}. These kinds of network use temporal data between frames to improve object classification. They accumulate information over a sequence of frames and learn patterns between time steps that may be important for the given label. So instead of treating each frame as a standalone feature, these networks relate new frames to previous frames often using memory mechanisms such as recurrent layers or temporal attention~\cite{peng2018two,qiao2022c3d}.

\section{Methodology}

\begin{figure*}[htp]
  \centering
  \includegraphics[width=0.8\textwidth]{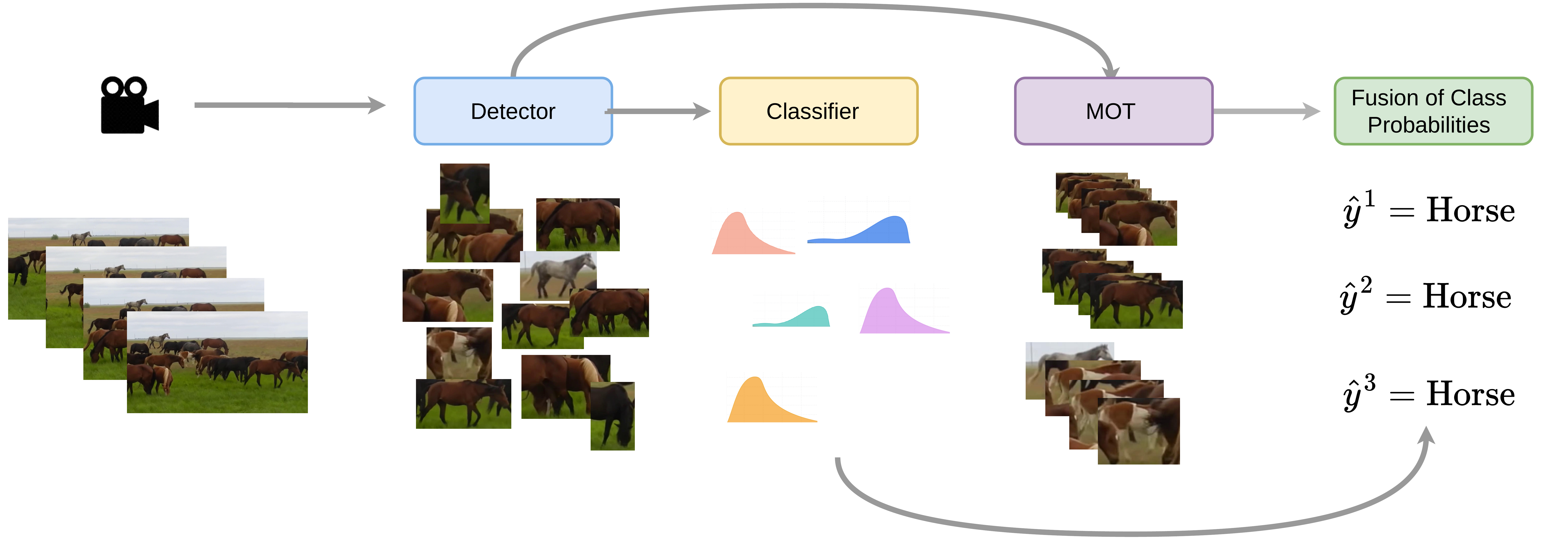} 
  \caption{The proposed framework processes sequential camera trap frames through a detector and a standalone classifier. The \gls{mot} module links these detections into individual trajectories, allowing for the Fusion of Class Probabilities to resolve label flickering.}\label{figure:framework}
\end{figure*}

\subsection{Wildlife Classification}

Wildlife species recognition is typically addressed as a multi-class image classification problem~\cite{clarfeld2025deepfaune}. Let $\mathcal{C} = \{c_1, c_2, \dots , c_{\vert \mathcal{C}} \vert \}$ denote a set of target class labels. Given an image of a animal $i^t$, captured as time $t$, a classifier based neural network $f_{\mathrm{cls}}$ produces a softmax-normalized scores $\mathbf{p}^t = f_{\mathrm{cls}}(i^t)$ over the probability distribution of the closed set of target labels such that $\sum_{n=1}^{\vert \mathcal{C} \vert} p^{t}_{n} = 1$. The softmax-normalized probabilities $p^{t}_{n} = P(c_{n} \vert i^{t})$ indicate the model's confidence that image $i^t$ contains the species $c_n$.

\subsection{Multi-Object Tracking}

\gls{mot} aims to track trajectories across a sequence of frames~\cite{liu2024deep}. Each trajectory traces the individual's unique identity. Let $\mathcal{I} = \{ i_1, i_2, \dots, i_{\vert \mathcal{I} \vert} \}$ be a set of identified objects. The number of visible identities may vary over time. Each frame in $\mathcal{T} = \{1, 2, \dots , T \}$ contains a subset of identities $\mathcal{I}_t \subseteq \mathcal{I}$ such that $t \in \mathcal{T}$. 
An object detection model $f_{\mathrm{det}}$ is applied to each frame to locate candidate objects. The detector outputs a set of bounding boxes $\mathcal{B} = \{ \mathbf{b}_{1}^{t}, \mathbf{b}_{2}^{t}, \dots\}$ such that each $\mathbf{b}_{k}^{t} \in \mathbb{R}^4$ encodes the spatial location of the detected object in image coordinates. However, the detector $f_{\mathrm{det}}$ does not provide the identity of each detected object.
\gls{mot} is tasked with solving the data association task. A \gls{mot} model $f_{\mathrm{mot}}$ determines which detections in frame $t$ correspond to previously observed detections in frame $t-1$. Formally, $f_{\mathrm{mot}}$ estimates a set of trajectory;
\begin{equation}
    \mathcal{O} = \{ O_1, O_2, \dots\}
\end{equation}
where each trajectory is defined as $O_k = \{ \mathbf{b}_{k}^{t} \vert t \in \mathcal{T}_{k} \}$ and $\mathcal{T}_{k} \subseteq \mathcal{T}$ represents the set of frames where $i_k$ is visible. The objective is to assign a unique label identity to each detection along its trajectory $\mathcal{O}_k$.

\subsection{Fusion of Class Probabilities}

Assuming that the correct association of objects has been done. We consider a single tracked object in $\mathcal{O}_k$ and describe the temporal fusion of class probabilities. Let 
\begin{equation}
    \mathbf{p}^{t-1}_{k} = \bigg( p^{t-1}_{k, 1}, \dots , p^{t-1}_{k, C} \bigg)
\end{equation}
denote the estimated class probability distribution for an identity with bounding box $\mathbf{b}_{k}^{t-1}$ at frame $t-1$, such that $p^{t-1}_{k, c} = P(c \vert \mathbf{b}_{k}^{t-1})$ is obtained from the classifier $f_{\mathrm{cls}}$. Given the next frame $t$, the classifier provides a new distribution $\mathbf{p}^{t}_{k}$ for an object with bounding box $\mathbf{b}_{k}^{t} \in \mathcal{O}_k$ within the same trajectory as $\mathbf{b}_{k}^{t-1}$. 
We enrich the prediction $\mathbf{p}^{t}_{k}$ with the context of previous frames, such that the augmented prediction becomes
\begin{equation}
    \hat{p}^{t}_{k, c} = \frac{p^{t-1}_{k,c} p^{t}_{k,c}}{\sum_{m=1}^{C}p^{t-1}_{k,c} p^{t}_{k,c}} 
\end{equation}
which we refer to as the \textit{fusion of class probabilities} using multi-object tracking. We assume that observations are conditionally independent given the true class.
To improve numerical stability and avoid underflow when multiplying small probabilities, we perform the update using logarithmic probabilities such that
\begin{equation}
    \hat{p}^{t}_{k, c} = \log p^{t-1}_{k,c} + \log p^{t}_{k,c} - \log \bigg( \sum_{m=1}^{\vert C \vert} \exp \big(\log p^{t-1}_{k,c} + \log p^{t}_{k,c}) \bigg)
\end{equation}
however, explicit normalisation can be omitted since only the maximum probability is required, hence
\begin{equation}
    \arg\max \hat{p}^{t}_{k, c} = \arg\max \bigg( \log p^{t-1}_{k,c} + \log p^{t}_{k,c} \bigg)
\end{equation}
is the updated class label for all objects tracked within the trajectory $\mathcal{O}_k$ up until time $t$.

\subsection{Camera Trap Simulation}

Our work aims to emulate realistic ecological monitoring conditions. We simulate motion-triggered camera trap data from continuous video datasets. Camera traps capture short bursts of images when motion is detected, followed by periods of inactivity. Given a video with sequence $\mathcal{T}$ frames, we simulate a motion-triggered event. When an object first enters the scene, a short burst of frames is initiated. The simulated camera trap captures a fixed number of frames, typically at one frame per second. Practitioners often collect one to two images per instance. This work will make use of four short bursts of frames, each one second apart.
If a given continuous source video dataset has a frame rate of $F$ frames per second, we subsample frames such that
\begin{equation}
    B(t_{m}) = \{ t_{m}, t_{m} + F, t_{m} + 2F, t_{m} + 3F\} \subseteq \mathcal{T}
\end{equation}
resulting in four sampled frames that follow temporally from each other.
Often, the camera trap has a short cool-down period after capturing a burst of frames. Typically, the cool-down period is five to ten seconds long. The study simulates a cool-down timer by skipping $\tau = 10$ seconds such that 
\begin{equation}
    t_{m+1} = t_{m} + \tau F
\end{equation}
represents the next burst of frames. 
The final objective of our framework is to produce a fused prediction
\begin{equation}
    \hat{y}^{m}_{k} = \arg\max_{c \in C} \sum_{t \in B(t_{m})} \log p^{t}_{k,c}
\end{equation}
for object $k$. Figure~\ref{figure:framework} depicts an illustration of our proposed framework.

\section{Experimental Setup}

\subsection{Implementation Details}

We adopt the standalone classifier from~\cite{muthivhi2025improving, huo2025nearest}; the authors use a pretrained backbone encoder from BioclipV2 and a two-layer classification head~\cite{gu2025bioclip}. We also use SAM3 as our object detection model to derive bounding boxes around detected animals~\cite{carion2025sam}. SAM3 is a prompt-based segmentation model. We set the prompt to \textit{animal} for each image and task it with finding the exact bounding box co-ordinates for each animal. Some \gls{mot} models may make use of a re-identification model to enhance their tracking ability. We use a pre-trained self-supervised re-identification model, developed using the strategy from \citet{muthivhi2025wildlife}, to extract fine-grained animal features~\cite{vcermak2024wildlifedatasets}.

\subsection{Baselines}
We adopt a variety of standard multi-target tracking methods to improve the classification performance of the standard classifier across a sequence of frames.
We used Intersection over Union (IoU)~\cite{Rezatofighi_2018_CVPR}, Simple Online and Realtime Tracking (SORT)~\cite{Bewley2016_sort}, Centroid~\cite{Nascimento1999AnAF}, Centroid with Kalman Filter~\cite{kalman1960new}, ByteTrack~\cite{zhang2022bytetrack}, BoostTrack~\cite{stanojevic2024boostTrack}, and finally BotSORT~\cite{aharon2022botsort}.

\subsection{Datasets}
The study uses three wildlife \gls{mot} datasets. \textbf{AnimalTrack} is a dedicated benchmark specifically designed for multi-animal tracking in the wild. The dataset consists of 58 video sequences that cover 10 common animal categories, with an average of 33 target objects per sequence. To ensure high-quality data for training and evaluation, every frame in the dataset has been manually labeled~\cite{zhang2023animaltrack}. \textbf{MammAlps} is a multimodal, multi-view dataset focused on wildlife behaviour monitoring. Collected using camera traps in the Swiss National Park, it contains over 14 hours of video with audio, 2D segmentation maps, and 8.5 hours of densely labelled individual tracks. The annotations cover 5 different species, 11 unique activities, and 19 unique actions~\cite{gabeff2025mammalps}. \textbf{SA-FARI} is a massive, large-scale dataset by Facebook built for multi-animal tracking and segmentation, comprising 11,609 camera trap videos collected over 10 years from 741 locations across four continents. It contains approximately 46 hours of footage spanning 99 wild animal species, and is exhaustively annotated with bounding boxes, individual identities, and high-quality spatio-temporal segmentation masks. We use the test set variant of the SA-FARI dataset. We restrict evaluation to the subset of animal classes that overlap with the label set trained on the classifier for each dataset~\cite{muthivhi2025improving,huo2025nearest}. That choice enables a direct, like-for-like comparison across datasets within a consistent class space. All remaining classes in each dataset are excluded from evaluation.

% , MOTIP~\cite{MOTIP}, SAM2MOT~\cite{jiang2025sam2mot} and finally SAM3~\cite{carion2025sam3segmentconcepts}.

\subsection{Evalution Metrics}

We use a standard multi-class classification metric to determine if a species has been successfully recognised across the datasets. \textit{Accuracy@1} (Acc@1) determines if the predicted label matches its ground truth annotation. The \textit{F1-Score} measures the harmonic mean between precision and recall. It penalises extreme values from both precision and recall. We report both the macro (F1-M) and weighted-average F1 scores (F1-W). 

%Macro-F1 calculates the F1-Score for each species and then averages the values. This process results in equal weighting of each species to the final score. Weighted F1-Score calculates the mean by weighting each species' F1-score by its relative prevalence in the ground truth data.

\section{Results}

\subsection{Classification}

\begin{figure*}[htp]
  \centering
  \includegraphics[width=\textwidth]{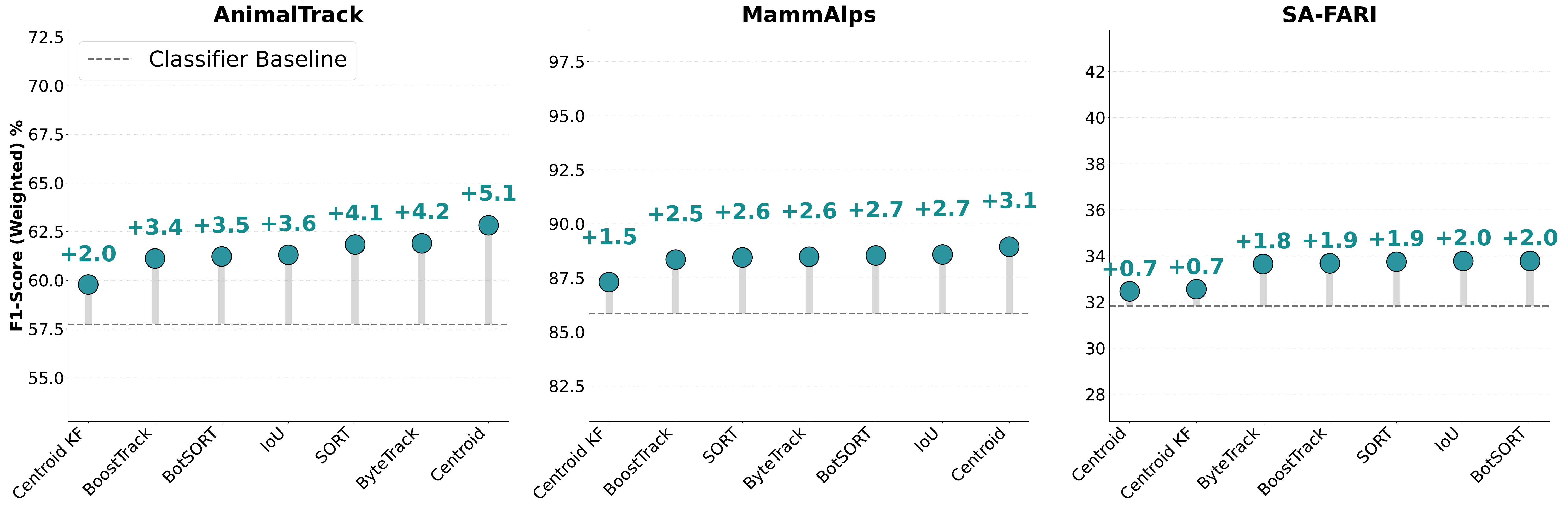} 
  \caption{The figure presents F1-score improvements across the AnimalTrack, MammAlps and SA-FARI datasets. The dotted line presents the classifier baseline performance and the green circles represents the improvement gains using the various proposed augmentation inference techniques.}\label{figure:dataset_gains}
\end{figure*}

\begin{table}[htbp]
\caption{Inference Augmentation results presented over the  AnimalTrack dataset}
\begin{center}
\resizebox{\columnwidth}{!}{%
\begin{tabular}{lrrrrrr}
\toprule
 \textbf{Method}
& Acc@1 & \makecell[l]{F1-M} & \makecell[l]{F1-W} \\
 \toprule
Classifier  & 45.20 & 11.89 & 57.75  \\
\textbf{\textit{Augmented Predictions}} \\
IoU         & 49.23 & 14.43 & 61.31  \\
SORT        & 49.90 & 14.55 & 61.84  \\
Centroid    & \ccB{51.44} & \ccB{16.85} & \ccB{62.83}  \\
Centroid KF & 47.28 & 12.72 & 59.78  \\
ByteTrack   & \ccb{49.70} & \ccb{14.58} & \ccb{61.90}  \\
BoostTrack  & 48.82 & 13.33 & 61.13  \\
% StrongSort  & 0.00 & 0.0 & 0.0 \\
BotSORT     & 49.09 & 13.79 & 61.23 \\
\bottomrule
\multicolumn{6}{l}{Top two along the columns are highlighted. Best model in bold}
\end{tabular}
}
\label{table:AnimalTack}
\end{center}
\end{table}

\begin{table}[htbp]
\caption{Inference Augmentation results presented over the MammAlps dataset}
\begin{center}
\resizebox{\columnwidth}{!}{%
\begin{tabular}{lrrrrrr}
\toprule
 \textbf{Method}
& Acc@1 & \makecell[l]{F1-M} & \makecell[l]{F1-W} \\
 \toprule
Classifier  & 77.16 & 6.06 & 85.86  \\
\textbf{\textit{Augmented Predictions}} \\
IoU         & \ccb{81.26} & 7.03 & \ccb{88.59}  \\
SORT        & 81.07 & 6.94 & 88.45  \\
Centroid    & \ccB{81.86} & \ccB{7.45} & \ccB{88.94}  \\
Centroid KF & 79.37 & 6.56 & 87.31  \\
ByteTrack   & 81.11 & 7.03 & 88.48  \\
BoostTrack  & 80.90 & 6.89 & 88.35  \\
% StrongSort  & 0.00 & 0.0 & 0.0 \\
BotSORT     & 81.19 & \ccb{7.10} & 88.54 \\
\bottomrule
\multicolumn{6}{l}{Top two along the columns are highlighted. Best model in bold}
\end{tabular}
}
\label{table:MAMMS}
\end{center}
\end{table}

\begin{table}[htbp]
\caption{SA-FARI}
\begin{center}
\resizebox{\columnwidth}{!}{%
\begin{tabular}{lrrrrrr}
\toprule
 \textbf{Method}
& \rot{Accuracy@1} & \rot{\makecell[l]{F1-Score\\(Macro)}} & \rot{\makecell[l]{F1-Score\\(Weighted)}} \\
 \toprule
Classifier  & 22.29 & 15.89 & 31.81  \\
\textbf{\textit{Augmented Predictions}} \\
IoU         & \ccB{24.34} & \ccb{20.64} & \ccB{33.78}  \\
SORT        & 24.34 & 20.60 & \ccb{33.75}  \\
Centroid    & \ccb{23.46} & \ccB{20.89} & 32.47  \\
Centroid KF & 23.46 & 18.63 & 32.56  \\
ByteTrack   & 24.34 & 20.51 & 33.65  \\
BoostTrack  & 24.34 & 20.54 & 33.68  \\
% StrongSort  & 0.00 & 0.0 & 0.0 \\
BotSORT     & \ccB{24.34} & \ccb{20.64} & \ccB{33.78} \\
\bottomrule
\multicolumn{6}{l}{Top two along the columns are highlighted. Best model in bold}
\end{tabular}
}
\label{table:SAFARI}
\end{center}

\end{table}

Tables~\ref{table:AnimalTack}, \ref{table:MAMMS}, and \ref{table:SAFARI} present Accuracy@1 (Acc@1), F1-score macro (F1-M) and F1-score weighted (F1-W) for the standalone classifier and our proposed inference-time augmentation. Performance improves across all augmentation variants on all three wildlife datasets. These results indicate that leveraging temporal context from consecutive frames and associating targets across frames can correct intermediate predictions and reduce frame-to-frame inconsistency. Figure~\ref{figure:dataset_gains} further summarises the F1-score gains of the augmented inference methods relative to the standalone baseline. The best-performing models, Centroid and BotSort, achieve the largest improvements of 5\%, 3.1\% and 2\% over the standalone classifier on the AnimalTrack, MammAlps and SA-FARI datasets.

\subsection{Per-Animal Performance}

\begin{figure*}[htp]
  \centering
  \includegraphics[width=\textwidth]{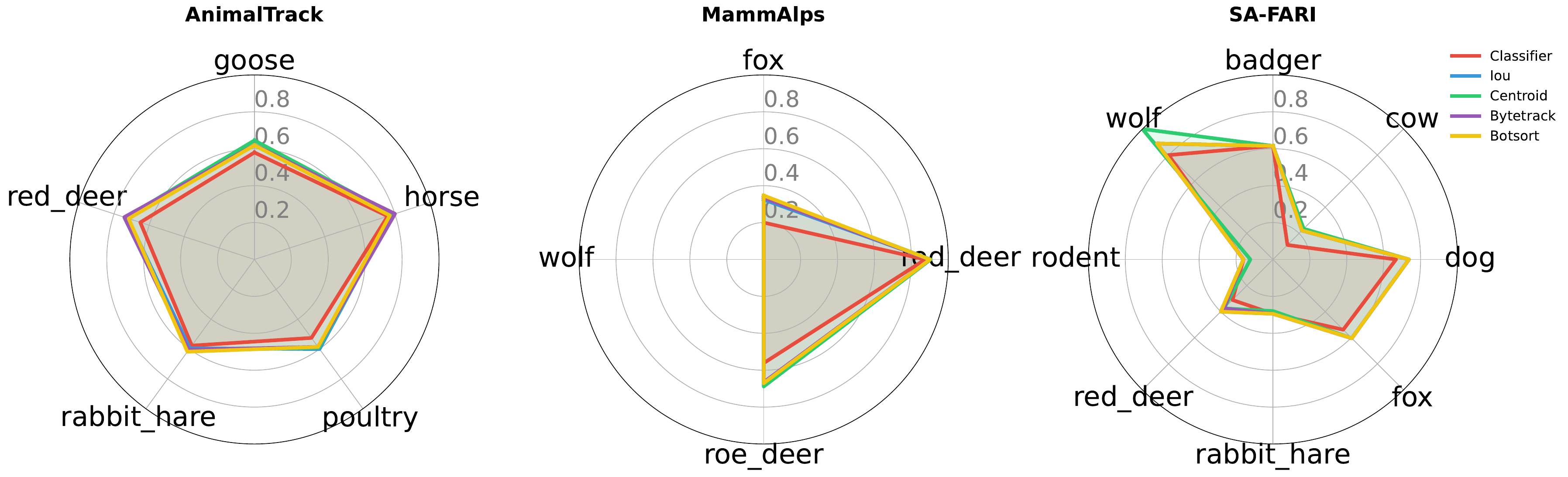} 
  \caption{The radar charts presents per animal class accuracy@1 performance on the classifier baseline and the improvement gains from the inference augmentation methods.}\label{figure:radar}
\end{figure*}

Figure~\ref{figure:radar} visualises per-class Accuracy@1 across three datasets for the standalone classifier and the proposed inference-time augmentations. Each axis corresponds to a class, and larger polygons indicate better performance. Across datasets, the augmented methods generally expand the polygon relative to the baseline, indicating consistent per-class gains rather than improvements driven by a single class. The magnitude of the gains varies by class and dataset, suggesting that temporal association is particularly helpful for classes where single-frame predictions are less stable across consecutive frames.

\subsection{Ablation}

\begin{table*}[htbp]
\caption{Ablation results assessing a Majority Voting strategy against our Probability Fusion}
\label{tab:ablation}
\begin{center}
\resizebox{\textwidth}{!}{%
\begin{tabular}{lrrr|rrr|rrrr|rr}
\toprule
 \textbf{Method} &
 \multicolumn{3}{c}{\textbf{AnimalTrack}} & 
 \multicolumn{3}{c}{\textbf{MammAlps}} & 
 \multicolumn{3}{c}{\textbf{SA-FARI}} \\
 \cmidrule(lr){2-4} \cmidrule(lr){5-7} \cmidrule(lr){8-10} 
& \makecell[l]{Acc@1} & \makecell[l]{F1-M} & \makecell[l]{F1-W} & \makecell[l]{Acc@1} & \makecell[l]{F1-M} & \makecell[l]{F1-W}  & \makecell[l]{Acc@1} & \makecell[l]{F1-M} & \makecell[l]{F1-W} \\
 \toprule
Classifier  & 45.20 & 11.89 & 57.75 & 77.16 & 6.06 & 85.86 & 22.29 & 15.89 & 31.81\\
\multicolumn{2}{l}{\textbf{\textit{Majority Voting}}} \\
Centroid    & 48.76 & 13.22 & 60.94 & 79.37 & 6.65 & 87.35 & 23.17 & 16.45 & 32.92 \\
BoostTrack  & 46.41 & 12.58 & 58.82 & 78.86 & 6.52 & 87.01 & 22.29 & 15.89 & 31.81 \\
BotSORT     & 47.15 & 12.77 & 59.55 & 79.03 & 6.51 & 87.11 & 23.46 & 16.48 & 33.33\\

\multicolumn{2}{l}{\textbf{\textit{Probability Fusion}}} \\
Centroid    & \ccB{51.44} & \ccB{16.85} & \ccB{62.83} & \ccB{81.86} & \ccB{7.45} & \ccB{88.94} & 23.46 & \ccb{20.89} & 32.47\\
BoostTrack  & 48.82 & 13.33 & 61.13 & 80.90 & 6.89 & 88.35 & \ccb{24.34} & 20.54 & \ccb{33.68} \\
BotSORT     & \ccb{49.09} & \ccb{13.79} & \ccb{61.23} & \ccb{81.19} & \ccb{7.10} & \ccb{88.54} & \ccB{24.34} & \ccB{20.64} & \ccB{33.78} \\

\bottomrule
\multicolumn{6}{l}{Top two along the columns are highlighted. Best model in bold}
\end{tabular}
}
\label{table:combined}
\end{center}
\end{table*}

Table~\ref{tab:ablation} quantifies the incremental benefit of adding each proposed component to a standalone classifier. Across MOT variants, linking detections across frames and applying majority vote yields consistent improvements of a few percentage points over the classifier alone, indicating that temporal association is already effective. Adding fusion of probability on top of MOT provides additional gains, showing that using the full confidence distribution further enriches the predictions rather than merely smoothing labels.

\subsection{Inference Time}
\begin{table}[htbp!]
\caption{Inference Time (milliseconds) per image sample}
\label{table:infertime}
\begin{center}
\resizebox{\columnwidth}{!}{%
\begin{tabular}{lr|rrrr}
\toprule
 \textbf{Method}
& \textbf{Total} & \makecell[l]{MOT} & \makecell[l]{ReID} & \makecell[l]{Classification} & \makecell[l]{Detection} \\
 \toprule
Classifier  & 1212.04 & - & - & +38.99 & +1173.04  \\
\textbf{\textit{Augmented Predictions}} \\
IoU         & 1213.07  & +1.02 & - & " & "  \\
SORT        & 1220.11 & +8.07 & - & " & " \\
Centroid    & 1213.25  & +1.21 & - & " & " \\
Centroid KF & 1213.61  & +1.57 & - & " & "  \\
ByteTrack   & 1236.59  & +22.52 & 2.01 & " & "  \\
BoostTrack  & 1246.27  & +32.21 & " & " & " \\
% StrongSort  &  &  & " & - & " & " \\
BotSORT     & 1246.63 & +32.56 & " & " & " \\
\bottomrule
%\multicolumn{6}{l}{Top two along the columns are highlighted. Best model in bold}
\end{tabular}
}
\label{table:infernce}
\end{center}
\end{table}

Table~\ref{table:infertime} reports per-sample inference time in milliseconds (ms) for the baseline, and each proposed augmentation method, measured on an NVIDIA RTX 5070 TI GPU. The table breaks down runtime into detection, classification, and additional tracking overhead (including ReID when applicable). Detection dominates the overall runtime, while classification contributes a comparatively small fraction. Lightweight association methods (IoU and centroid) add only minor overhead. ReID-based and stronger trackers incur additional cost due to appearance matching. Overall, the added tracking time remains small relative to detection, keeping the augmented inference practical while improving prediction stability.

\subsection{Qualitative Analysis}

\begin{figure*}[htp]
  \centering
  \includegraphics[width=\textwidth]{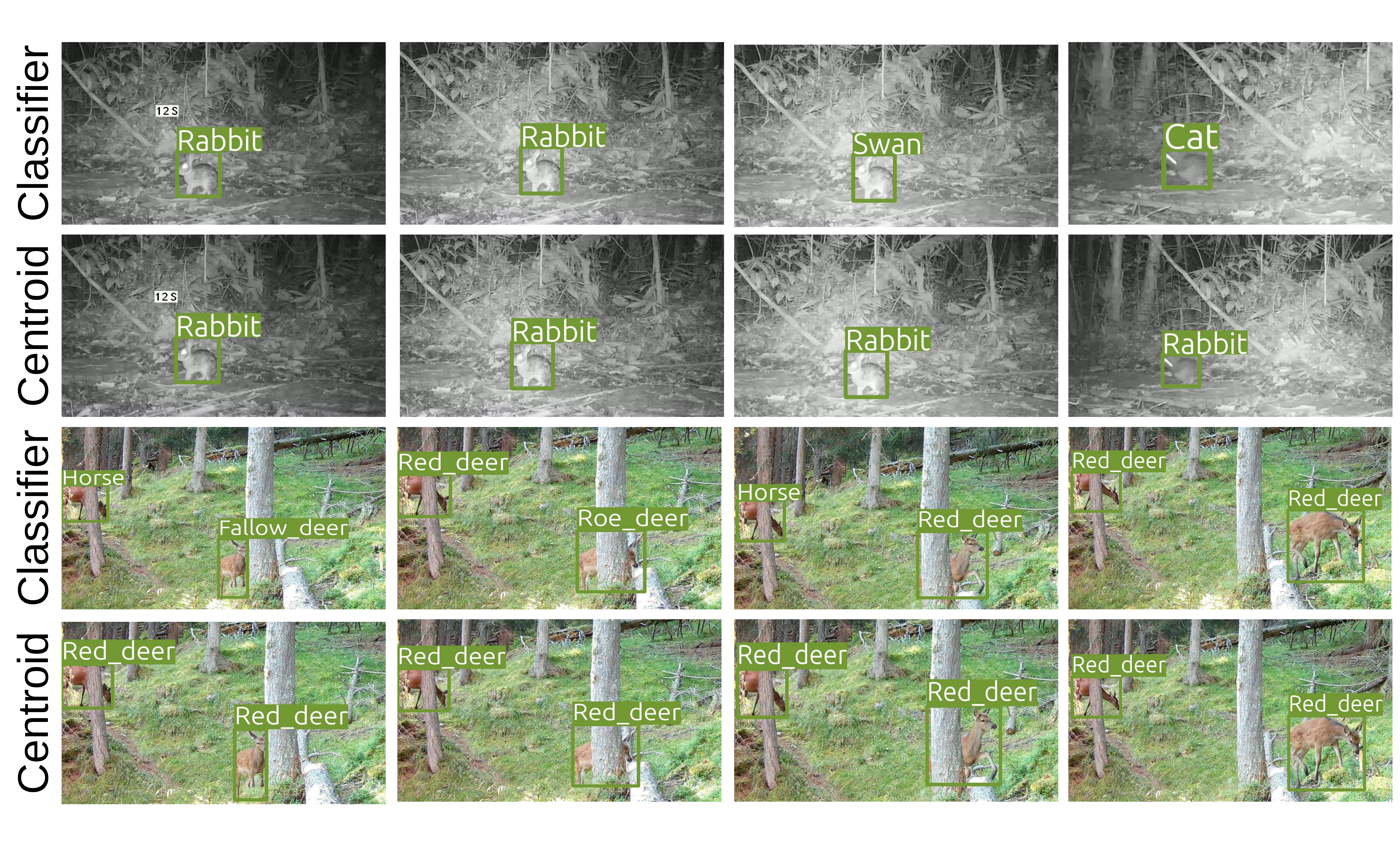} 
  \caption{Consecutive-frame examples comparing the standalone classifier (top) to centroid-based MOT (bottom) of each example. Centroid association reduces frame-to-frame label flicker and yields more consistent predictions for the same target. For instance (top row) the rabbit is initially miss-classified as a Swan and Cat; also, (third row) labels flicker between Horse, Roe, Fallow and Red Deer.}\label{figure:qualitative}
\end{figure*}
Figure~\ref{figure:qualitative} illustrates the effect of MOT association on prediction stability. The standalone classifier can change labels across consecutive frames for the same animal.  In the night sequence, the predicted label briefly switches away from \emph{Rabbit}. In the daytime sequence, the deer label also varies from frame to frame. 
Centroid-based association links detections across frames and enforces temporal consistency. Incorrect intermediate predictions are therefore suppressed, leading to more stable sequence-level output.

\section{Conclusion}
\subsection{Broader impact}
Our proposed inference-time augmentation consistently improves wildlife classification across AnimalTrack, MammAlps, and SA-FARI without retraining the classifier. All multi-object tracking (MOT) variants outperform the standalone baseline, and the strongest methods (Centroid and BotSORT) achieve the highest macro F1 gains (up to 5\%, 3.1\% and 2\% on AnimalTrack, MammAlps and SA-FARI, respectively). Per-class trends indicate that improvements are broadly distributed rather than driven by a single dominant class. Qualitative results further provide clear evidence of reduced frame-to-frame label changes. Practical deployment is supported by the runtime breakdown: detection dominates total cost, and lightweight association adds only minor overhead, keeping the approach feasible for monitoring pipelines that require stable predictions at scale.

\subsection{Limitations and Future Work}
Several aspects remain important to address. The approach relies on reliable detections and cross-frame association. Missed detections, fragmented tracks, or identity switches can still cause fusion to combine inconsistent predictions. The benefit also depends on the available temporal context, so very short sequences or brief appearances provide fewer frames to stabilise predictions. Class imbalance and systematic biases in the underlying classifier can be reinforced when probabilities are accumulated across frames, particularly for rare species. The evaluation reports frame-level Accuracy@1 and F1. These metrics capture classification quality well, but they do not directly quantify label flicker. Adding temporal stability metrics would better align the evaluation with the goal of this work. Suitable measures include label flip rate and track-level classification accuracy. Future work can also strengthen the robustness of the association, refine fusion strategies, and expand on sequence-aware evaluation.

\bibliographystyle{plainnat}
\bibliography{bibliography}

% \bibitem{b1} G. Eason, B. Noble, and I. N. Sneddon, ``On certain integrals of Lipschitz-Hankel type involving products of Bessel functions,'' Phil. Trans. Roy. Soc. London, vol. A247, pp. 529--551, April 1955.
% \bibitem{b2} J. Clerk Maxwell, A Treatise on Electricity and Magnetism, 3rd ed., vol. 2. Oxford: Clarendon, 1892, pp.68--73.
% \bibitem{b3} I. S. Jacobs and C. P. Bean, ``Fine particles, thin films and exchange anisotropy,'' in Magnetism, vol. III, G. T. Rado and H. Suhl, Eds. New York: Academic, 1963, pp. 271--350.
% \bibitem{b4} K. Elissa, ``Title of paper if known,'' unpublished.
% \bibitem{b5} R. Nicole, ``Title of paper with only first word capitalized,'' J. Name Stand. Abbrev., in press.
% \bibitem{b6} Y. Yorozu, M. Hirano, K. Oka, and Y. Tagawa, ``Electron spectroscopy studies on magneto-optical media and plastic substrate interface,'' IEEE Transl. J. Magn. Japan, vol. 2, pp. 740--741, August 1987 [Digests 9th Annual Conf. Magnetics Japan, p. 301, 1982].
% \bibitem{b7} M. Young, The Technical Writer's Handbook. Mill Valley, CA: University Science, 1989.
% \end{thebibliography}
% \vspace{12pt}
% \color{red}
% IEEE conference templates contain guidance text for composing and formatting conference papers. Please ensure that all template text is removed from your conference paper prior to submission to the conference. Failure to remove the template text from your paper may result in your paper not being published.

\end{document}